\documentclass{article} 
\usepackage{nips13submit_e,times}
\usepackage{hyperref}
\usepackage{graphicx,amsmath,amssymb,algorithm,algpseudocode}
\usepackage{url}

\title{Learning Deep Face Representation}

\author{
Haoqiang Fan \\
Megvii Inc.\\
\texttt{fhq@megvii.com} \\
\And
Zhimin Cao \\
Megvii Inc. \\
\texttt{czm@megvii.com} \\
\And
Yuning Jiang \\
Megvii Inc.\\
\texttt{jyn@megvii.com} \\
\And
Qi Yin\\
Megvii Inc. \\
\texttt{yq@megvii.com} \\
\And
Chinchilla Doudou \\
Megvii Inc. \\
\texttt{doudou@megvii.com} \\
}

\nipsfinalcopy 

\begin{document}

\maketitle

\begin{abstract}

Face representation is a crucial step of face recognition systems. An optimal face representation should be discriminative, robust, compact, and very easy-to-implement. While numerous hand-crafted and learning-based representations have been proposed, considerable room for improvement is still present.
In this paper, we present a very easy-to-implement deep learning framework for face representation. Our method bases on a new structure of deep network (called Pyramid CNN). The proposed Pyramid CNN adopts a greedy-filter-and-down-sample operation, which enables the training procedure to be very fast and computation-efficient. In addition, the structure of Pyramid CNN can naturally incorporate feature sharing across multi-scale face representations, increasing the discriminative ability of resulting representation. Our basic network is capable of achieving high recognition accuracy ($85.8\%$ on LFW benchmark) with only 8 dimension representation. When extended to feature-sharing Pyramid CNN, our system achieves the state-of-the-art performance ($97.3\%$) on LFW benchmark.
We also introduce a new benchmark of realistic face images on social network and validate our proposed representation has a good ability of generalization.

\end{abstract}

\section{Introduction}

The very first step in most face recognition systems is to represent the facial images as feature vectors. After obtaining the representation, various learning algorithms can be applied to perform the task of classification, verification or searching. Performance of the algorithms heavily depends on the choice of representation. For that reason, considerable efforts have been devoted to designing better representation based on the prior knowledge of face images. Despite the progress in this direction, the current state-of-the-art system on the LFW benchmark is built upon the hand-crafted LBP descriptor~\cite{ojala02} which was proposed more than a decade ago. Even worse, \cite{dong13}'s experiments showed that most hand-crafted features only gave similar results under the high-dimensional learning framework. It is arguably claimed that traditional hand-crafted representations suffered from a visible performance bottleneck and most of them were making different tradeoffs between discriminative ability and robustness.

\begin{figure}[t]
\centering
\includegraphics[width=0.9\linewidth]{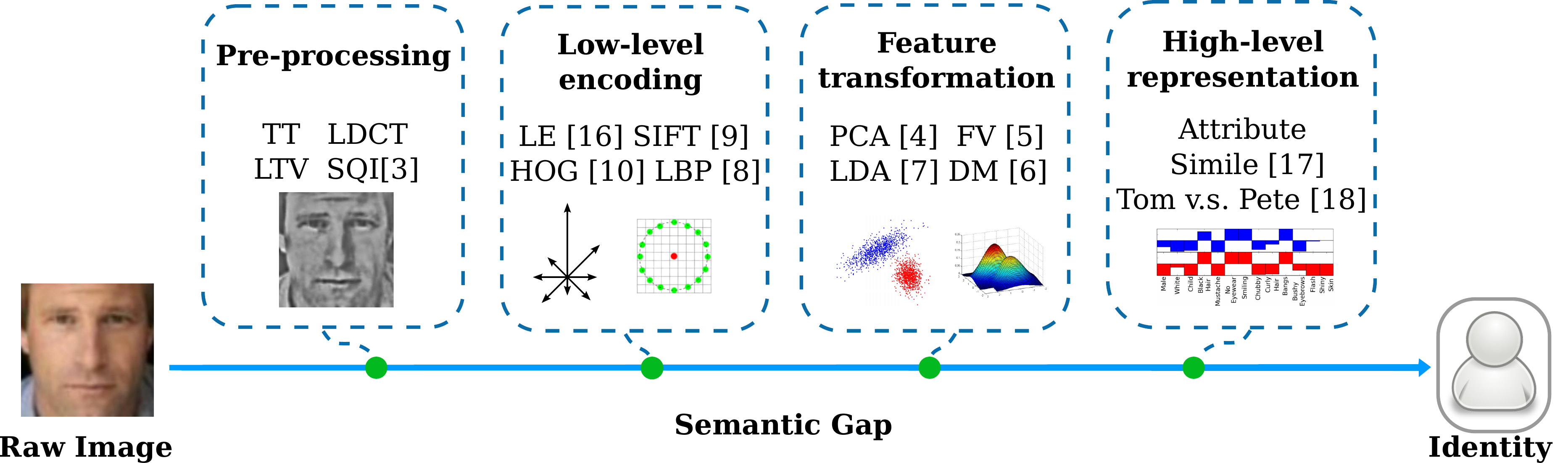}
\caption{The traditional face representation pipeline. The image is preprocessed, encoded and transformed to representations with various level of semantics. Existing methods focus on improving the building blocks in this pipeline to close the semantic gap between image pixels and the identity. The proposed method is to unify the pipeline with deep learning techniques. The neural network directly works on the image pixels. Inside the network, the signal undergoes multiple layers of non-linear transformations which functionally resemble the manually designed multi-stage pipeline.}
\label{fig:pipeline}
\end{figure}

To upgrade the performance, numerous learning algorithms are employed to infer better face representation. However, most of these methods are complicated multi-stage systems and optimized separately for each module. Figure~\ref{fig:pipeline} lists some typical methods used in each stage. The pipeline consists of pre-processing~\cite{han13}, low-level encoding, feature transformation~\cite{sirovich87,simonyan13,barkan13,fisher36} and higher level representation. The careful tuning of each individual module is very labor-intensive. More important, it is unclear how to ensure the performance of the whole system by optimizing each module individually.

In this paper, we present a unified and very easy-to-implement framework for face representation learning. 
Our method bases on a new structure of deep convolutional neural network (called Pyramid CNN) with supervised learning signals in the form of face pairs.
Our network extracts features directly from image pixel and outputs a highly-compact representation after training of a huge number of face images.
With a specially designed operation of filter-and-down-sample, our network can be trained very fast and computation-efficient and achieve high 
recognition accuracy with an extreme compact feature (only 8 dimensions).

In addition, the structure of our Pyramid CNN can naturally incorporate feature sharing across multi-scale face representations,
increasing the discriminative ability of resulting representation significantly.
With our multi-scale feature sharing network, we achieve the state-of-the-art performance ($97.3\%$) on LFW benchmark.
To validate the generalization ability of our learned representation, we introduce a new face benchmark collected from realistic face images on social network.
Experiments shows that our method outperforms the traditional face representation methods significantly.

The contributions of this paper are summarized as follows:

\begin{enumerate}
\item We present a unified and very easy-to-implement framework for deep face representation learning.
\item We propose a new structure of network (Pyramid CNN), which can extract highly discriminative and compact representation and be trained in a fast and computation-efficient manner.
\item Our method can achieve high recognition performance with extremely compact feature (8-dimension). It is also able to achieve the state-of-art performance (97.3\%) on the LFW benchmark.
\item We introduce a new benchmark collected from realistic face images of social network and validate the generalization ability of our representation.
\end{enumerate}

\section{Related Work}

Numerous hand-crafted feature extractors have been used in face recognition. LBP~\cite{ojala02,ahonen04}, SIFT~\cite{lowe99}, HOG~\cite{dalal05}, Gabor~\cite{shen06} and other descriptors are proposed based on various heuristics. Some descriptors aim to be an extension or improvement of existing ones (e.g. LTP~\cite{tan07}, TPLBP, FPLBP~\cite{wolf08}, CLBP~\cite{guo10}, LGBPHS~\cite{zhang05}, LE~\cite{cao10}). Designing these hand-crafted and hard-wired descriptors is usually a trial-and-error process with careful tuning of the parameters. Also, these low level features only manifest semantic meaning when a large number of dimensions work together.

Effort is made to represent the face by higher level attributes. The straight forward way is to manually choose several attributes~\cite{kumar09} and build classifiers to model them. The classifier's output is then used as the face's representation. This method encounters difficulties when the dimension of the representation becomes high. Also, the performance relies on the particular choice of the set of attributes. This idea can be extended to use the similarity with reference person as the attribute~\cite{kumar09,berg12}. Unsupervised methods based on clustering are also proposed~\cite{barkan13}. These works are usually built on low level shallow features.

Deep learning techniques have been applied in face recognition. One direction is to automate the representation learning process by unsupervised learning algorithms \cite{huang12,nair10,ranzato11}. \cite{cox11} used screening to select the features from a large number of randomly generated candidates. Another direction is to directly learn the similarity score by deep neural networks~\cite{chopra05,sun13}. Our method is inspired by~\cite{chopra05}. However, substantial difference exists. Instead of using the CNN as a black box to match the face pairs, we use the CNN to learn a representation of the face that is utilized by recognition algorithms in latter stage. Also, we exploit the connection between the layerwise architecture of the CNN and the multi-scale structure of the facial image.

\section{Our Method}

\subsection{Considerations for the Design of Face Representation}

The representation is a function map from image pixel to numeric vector.
\[
f: \mathbb{R}^{h \times w} \rightarrow \mathbb{R}^{m}
\]

We state some natural criteria for a good face representation.

\textbf{Identity-preserving.} The distance in the mapped space should closely reflect the semantic distance of the identity of the face image. The influence of irrelevant factors should be minimized.

\textbf{Abstract and Compact.} The short length of the representation, which corresponds to small $m$, is especially appreciated by recognition models. To keep discriminative power in the low dimensional space, the representation should encode abstract and high level information of the identity of the face.

In addition to these criteria, we have another consideration for the representation's design procedure.

\textbf{Uniform and Automatic.} The hand-crafted and hard-wired part should be minimized to automate the design of representation. The ideal method is to close the semantic gap within a single uniform model.

One way to obtain the desired representation is to learn it from data. This involves parameterizing a function family and using an object function $L$ to choose the representation extractor.
\[
	\theta_0 = \underset{\theta}{\text{argmin}} ~ L(f_\theta, I_{\text{data}})
\]
The underlying function family $f$ should contain enough complexity to express the complex and high-level computation required. To guarantee the ID-preserving property, identity information should be used in the objective function $L$. This leads to a supervised representation learning method which is in contrast to the unsupervised methods that aim at modeling the density distribution of data points. Though unsupervised methods are capable of discovering the patterns emerged in data, their optimization target is not directly related to the recognition task, so the learned representation is inevitably influenced by factors including illumination, expression and pose.

The above considerations lead us to a supervised deep learning based representation learning method. The supervised signal explicitly impose the ID-preserving requirement. The deep neural network directly works from raw image pixels, and it is capable of expressing highly non-linear and abstract computation. We also propose a new method which exploits the property of the facial images to improve the training of deep neural networks.

\subsection{Pyramid CNN}

Our method is based on CNN. The neural networks are applied to image patches, and their last layer activation is taken as the representation. To train the CNN, supervised signal is used.

\textbf{ID-preserving Representation Learning.} The most abundant type of supervised signal in face recognition is face pairs labeled with whether they belong to the same person. Numerous matched and unmatched pairs can be generated from photo gallery. Moreover, the pairs can also be obtained from group photo or video.

We adopted the ``Siamese networks''~\cite{bromley93} to handle the face pairs. The same CNN is applied to the two images to produce the corresponding representation. One output neuron uses a distance function to compare the representation and predicts whether the face pair belong to the same person. The loss function is
\begin{equation} \label{eq:obj}
	L=\sum_{I_1,I_2} \log (1+\exp(\delta(I_1,I_2) D(I_1,I_2)))
\end{equation}
\begin{equation}
	D(I_1,I_2) = \alpha \cdot d(f_{\theta}(I_1),f_{\theta}(I_2)) - \beta
\end{equation}
where $\delta(I_1,I_2)$ indicates whether the two images $I_1$ and $I_2$ belong to the same person. $f_{\theta}$ represents the computation done by the neural network, and $d$ is a function to measure the distance between the two vectors. $\theta$ stands for the weights in the network. $\alpha$ and $\beta$ are trainable parameters.

The loss function encourages small distance between features belonging to the same person and penalizes the similarity between unmatched pairs. In this way, the learned feature manifests good ID-preserving property, and it is highly specific to the recognition task. Factors corresponding to intra-person variation will be suppressed by the network.

\begin{figure}
\centering
\includegraphics[height=5cm]{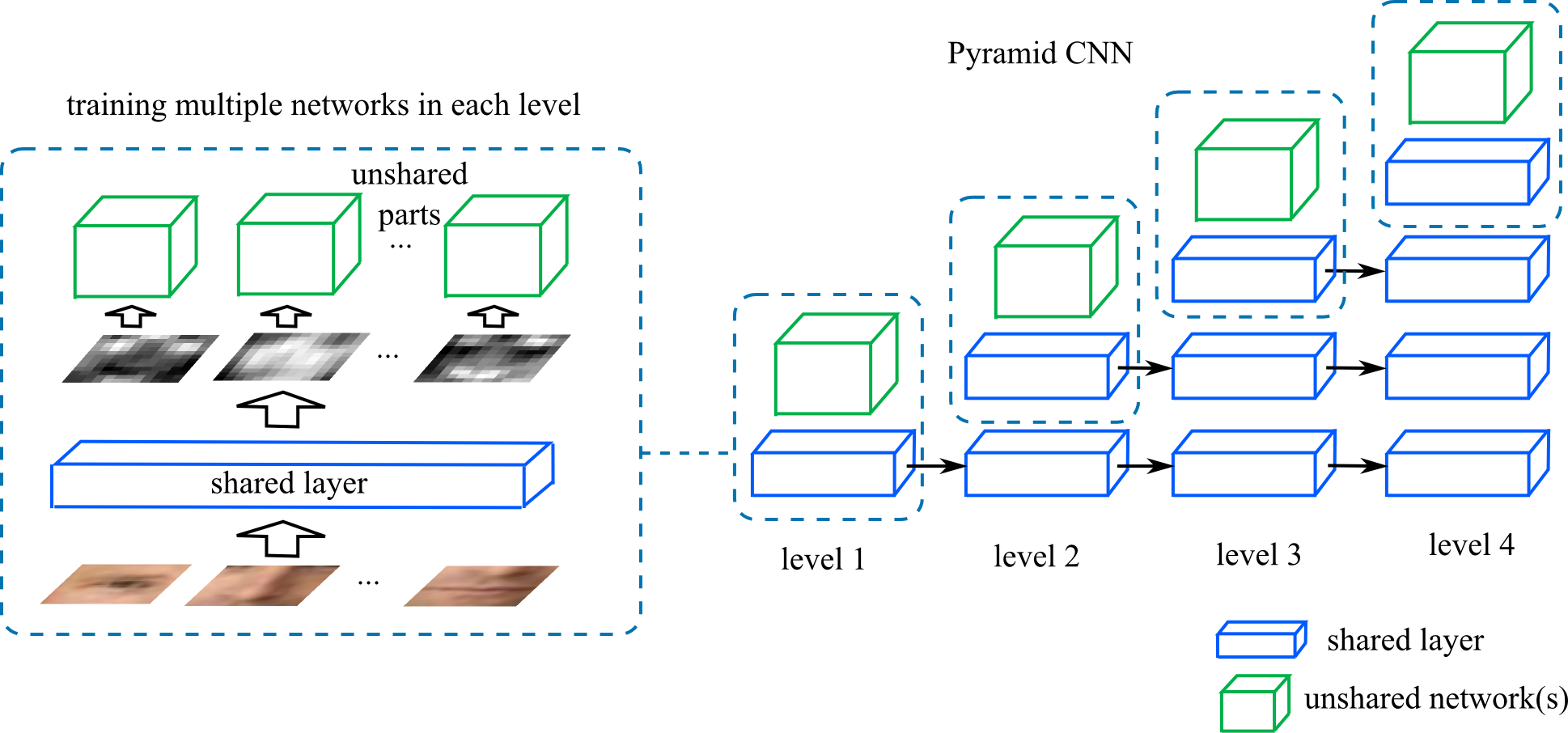}
\caption{ Our Pyramid CNN method. To train an individual network, the ``Siamese'' network is used. Two images are fed to the same CNN and the outputs are compared by the output neuron which predicts whether the two faces have the same identity. The Pyramid CNN consists of several levels of networks. These networks have different depths and input sizes, and they share some of the layers. The pyramid is trained in a greedy manner. The first network is trained on part of the face, and its first layer is fixed. The fixed layer is used to filter and down-sample input to higher level networks. Higher level networks are trained on the processed images. In this way, the size of the network that is actually trained does not grow as the level increases.
}
\label{fig:pyramid}
\end{figure}

\textbf{Pyramid CNN architecture.} A convolutional neural network is a highly non-linear real valued multivariate function which is composed of multiple layers of convolution and down sampling operators:
\begin{equation}
I^{i+1}=P_\text{max}(g(I^{i} \otimes W^{i} + B^{i}))
\end{equation}
where $I^{i}$ is the neuron's value in the $i$-th layer,  $g$ is a non-linear activation function. The convolution and down-sampling operator:
\begin{equation}
(I^{i} \otimes W^{i})_{x,y,z}=\sum_{a,b,c,z} I^{i}_{x-a,y-b,c} W^{i}_{a,b,c,z}
\end{equation}
\begin{equation}
P_\text{max}(I^{i})_{x,y}=\max_{0 \leq a,b < s} I^{i}_{xs + a, ys + b}
\end{equation}
where $W$ and $B$ are the network's weights.
 Gradient based algorithms are developed to estimate the weights in the network. Deeper and larger networks are generally more powerful. However, the time required to train a neural network grows rapidly as its size and depth is increased.

The motivation of our Pyramid CNN is to accelerate the training of deep neural networks and take advantage of the multi-scale structure of the face. Figure~\ref{fig:pyramid} gives a illustration of the structure of the Pyramid CNN. The Pyramid CNN is a group of CNNs divided into multiple levels. A network is composed of several shared layers and an unshared part which has the same structure at all levels. The first layer is shared across all levels. The second layer is shared by networks from the second level. This sharing scheme is repeated. Due to the down-sampling operation in the shared layers, the input size of the network at higher level is larger. It is allowed (though not shown in the figure) that more than one networks exist in the same level, and they work on different region while sharing their first layer parameters.

One view of the Pyramid CNN is a supervised layerwise training method for a deep neural network. The actual depth of the network in the last level is increased by the layers it shares with lower levels. However, it is not necessary to train that deep (and large) network directly. The training procedure can be decomposed into training several small networks.

\begin{algorithm}[h]
\begin{algorithmic}
\State Input: the image data $I_{\text{data}}$
\For{ $l$ from 0 to $l_{\text{max}}$}
	\State set $I_{\text{train}}$ to patches cropped from $I_{\text{data}}$
	\State let the CNN be $f_{\theta}(x) = f'_{\theta}(f^1_{\theta}(x))$, where $f^1$ is the first layer's computation.
	\State find $\theta$ by minimizing the objective (\ref{eq:obj}) with $I_{\text{data}}$
	\State process the image data $I_{\text{data}} \leftarrow f^1_{\theta}(I_{\text{data}})$
\EndFor
\end{algorithmic}
\caption{Supervised Greedy Training of the Pyramid CNN.}
\end{algorithm}

The network in the first level which has relatively small input size is first trained on part of the face. After training the first level network, its first layer is fixed and the fixed layer is used to process (filter and down-sample) the training images. The second level networks are trained on the processed image. In this way, the input size of the network that is actually trained does not become larger as the level increases. This greedy layerwise training procedure continues until networks at all levels are trained, and the final network with extra depth is obtained. The purpose of allowing more than one networks in the same level is to compensate for the lower level network's small coverage of the input region.

In deep learning literature, the pre-training technique is widely used. However, most pre-training algorithms are unsupervised the optimization target is loosely related to recognition. Our method is strongly supervised in that the learning signal at all levels directly reflect the final task. Thus, it is guaranteed that the shared layers learn to extract discriminative information closely related to the task.

Another interpretation of the Pyramid CNN is a multi-scale feature extraction architecture. The pyramid can naturally handle multi-scale input patch which is common in face recognition. The image patches of different sizes are fed to networks at corresponding scale level. The pyramid takes advantage of the multi-scale structure by using deeper networks for larger input region. The increase in depth allows higher level networks to undertake more complex and abstract computation on larger image patches.

\textbf{Landmark Based Pyramid CNN.} Previous work~\cite{dong13} showed that face verification algorithms are able to benefit from over-complete high dimensional representation. The feature vectors corresponding to different scale levels can be concatenated to form a long representation. Also, extracting the features based on detected landmark positions is important in improving the robustness of the system. To enable fair comparison with other methods, we extend our Pyramid CNN to the landmark based multi-scale feature extraction scheme. The pyramids are built on different landmark positions, and the output at all levels of the pyramids are concatenated to increase the dimension of the representation.

\section{Experiments}

\subsection{Results on LFW benchmark}

LFW is a challenging benchmark for face recognition systems. It contained more than 13,000 pictures acquired from the web. The experiment protocol is to evaluate the accuracy of verifying whether the two images belong to the same person. Our results are compared with other methods that also used outside training data.

\begin{figure}
\centering
\includegraphics[height=5.5cm]{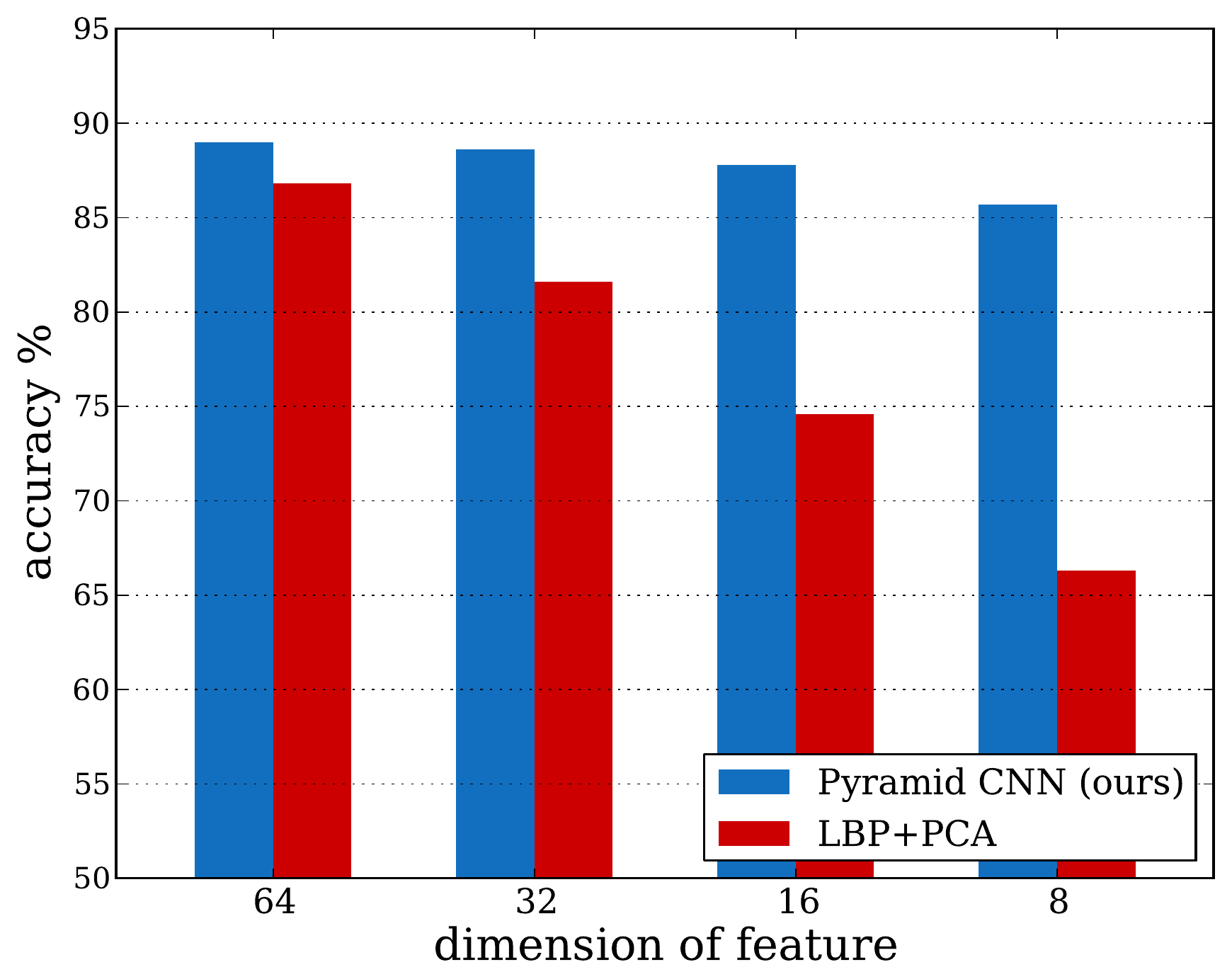}
\caption{
	The performance of the representation when the dimension is low. Our representation is the output of the highest level CNN in the pyramid. The baseline is the over-complete LBP feature reduced to various dimensions by PCA. Though the LBP baseline has an accuracy of 96.3\% when its dimension is above 1000, its performance is poor in the low dimensional setting. In contrast, the performance of our compact deep learning based representation only drops slowly when the dimension is reduced.
}
\label{fig:compact}
\end{figure}

\textbf{Compact Single Feature} We first run a 4 level Pyramid CNN on the whole face image. The output of the last level network is taken as the representation of the image, and other levels are only constructed for training. The representation is learned on an outside face album containing the photos of thousands of people, and it is tested on the LFW benchmark. Figure~\ref{fig:compact} shows the accuracy of this feature at different number of dimensions. Compared to the LBP baseline, the performance of our representation only deteriorates slowly as the dimension is reduced.

\begin{figure}
\centering
\includegraphics[height=7cm]{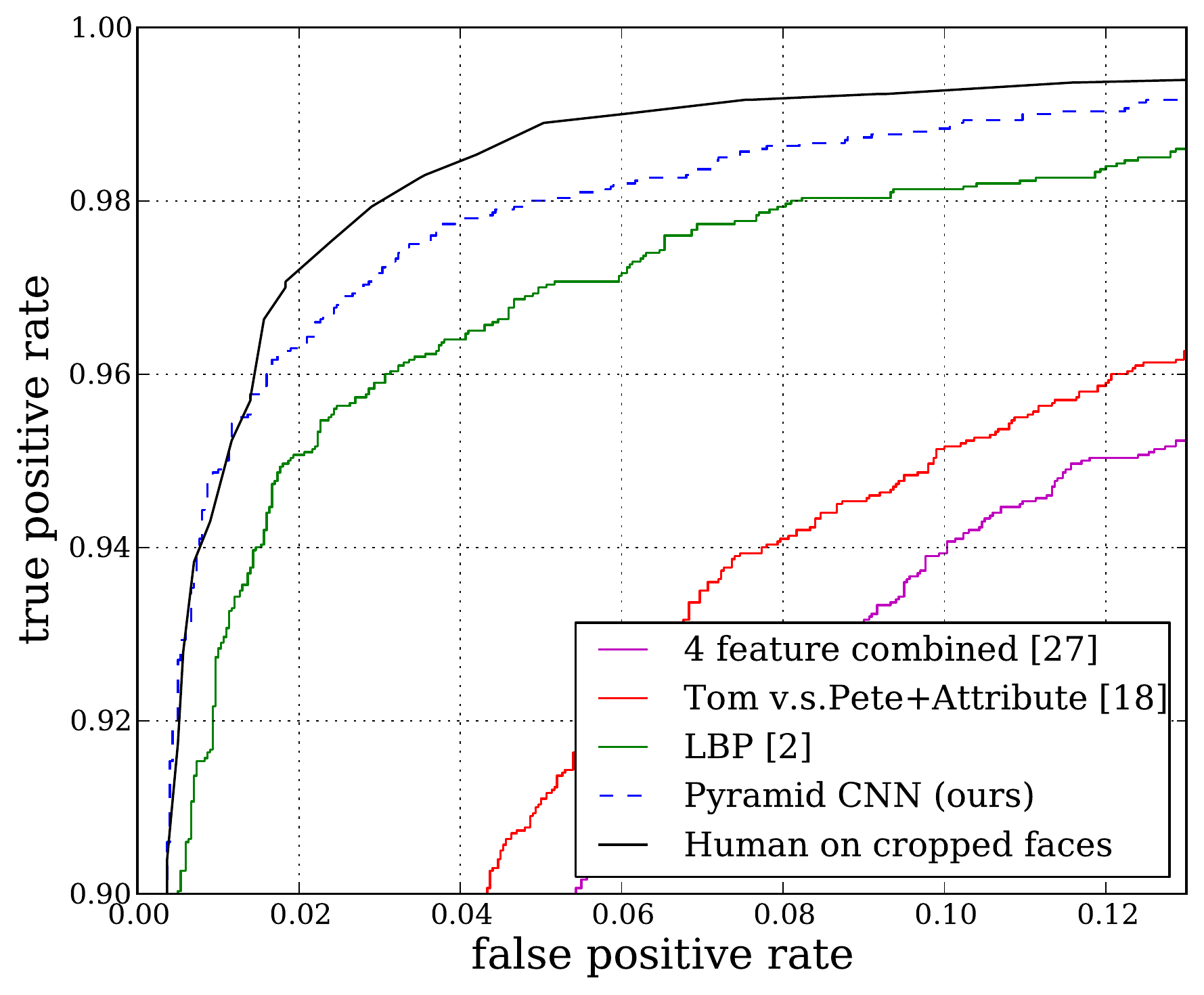}
\caption{
	Performance on the LFW benchmark when there is no constraint on the length of the representation. Our system achieves the state-of-the-art performance and it is very close to human performance on cropped images.
}
\label{fig:lfwroc}
\end{figure}

\textbf{High Dimensional Representation.} To fairly compare our representation method with other systems that enjoy the bless of dimensionality, we followed the common practice of extracting landmark based multi-scale representation. As shown in Figure~\ref{fig:lfwroc}, our system achieved state-of-the-art performance on the LFW benchmark. Furthermore, the accuracy is very close to human's performance when the face is cropped by a bounding box.

\begin{figure}
\centering
\includegraphics[height=7cm]{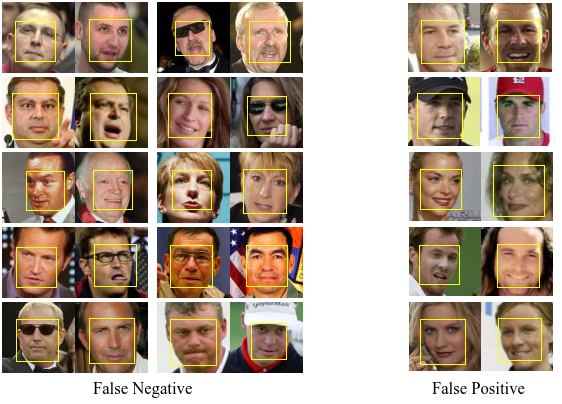}
\caption{
	Examples of the incorrect predictions on LFW view 2. We made totally 164 errors. 15 pairs are sampled in this figure. Due to the effect of aging and occlusion, these cases are not easy even to human. Green rectangle is for the pairs that the ground truth labels as not belonging to the same person. 
}
\label{fig:ourerrors}
\end{figure}

\textbf{Error Analysis.} Our system made 164 incorrect predictions. The ``ground truth'' label of 3 of them are wrong. Another 3 mistakes can be attributed to the face detector. Among the remaining 158 pairs, we sampled 15 pairs in Figure~\ref{fig:ourerrors}. As shown in the figure, these cases are not easy even for human due to the effect of aging, occlusion as well as other factors. We argue that further improvement will rely more on the context information in the image as well as the background knowledge about the famous people in this dataset.

\subsection{Effect of Layer Sharing}

\begin{figure}
\centering
\includegraphics[height=7cm]{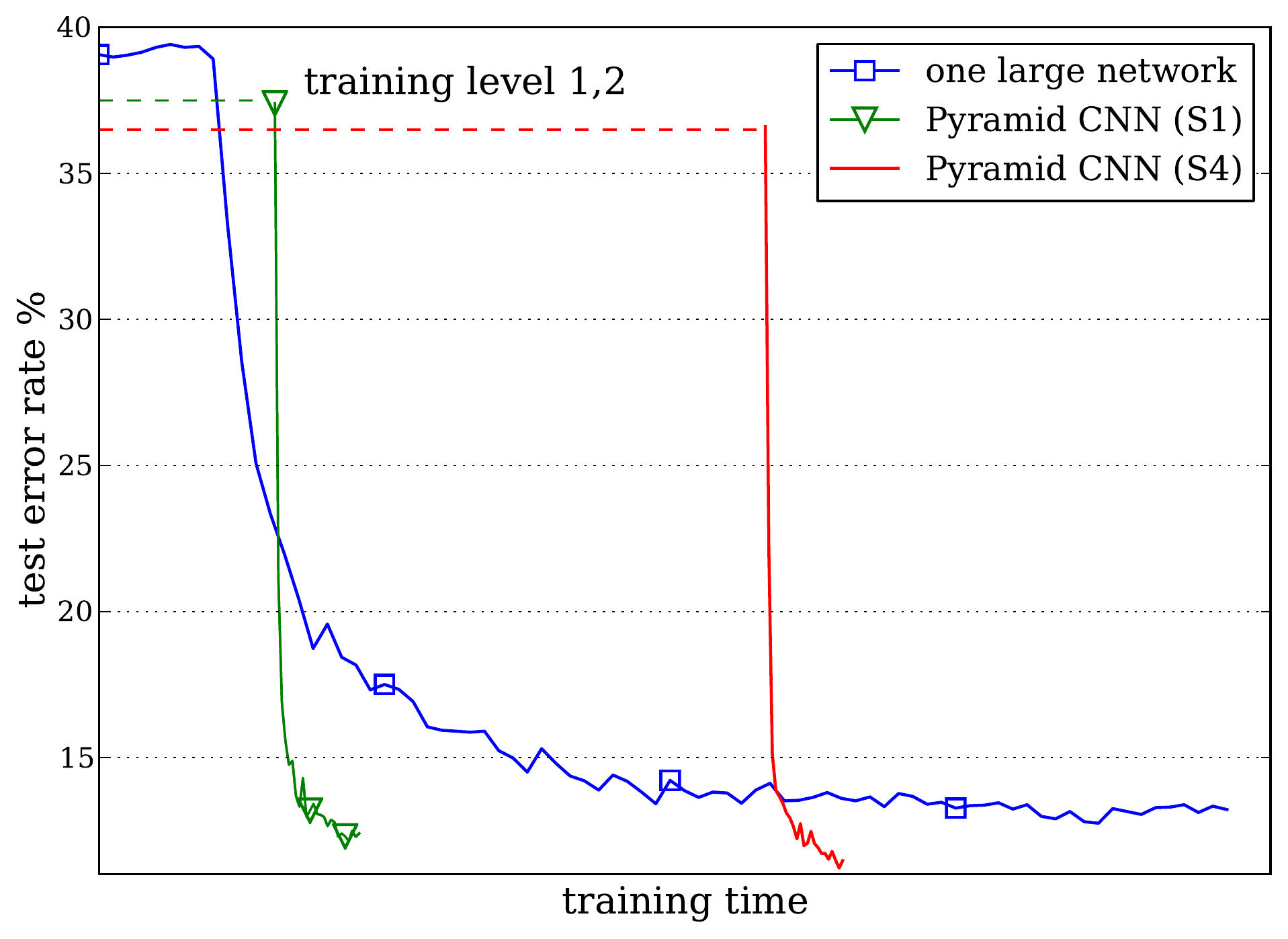}
\caption{
	Pyramid CNN accelerates training of large and deep CNN. The figure shows the test accuracy at different time points in training. Given the same amount of time, networks trained by Pyramid CNN achieve significantly better result. Also, using more networks in one scale level in the pyramid improves performance.
}
\label{fig:traintime}
\end{figure}

We conducted comparative experiments to study the effects of the layer sharing mechanism in the Pyramid CNN. In this experiment, we trained a 3 level Pyramid CNN, with one or four networks in the same scale level. Then we trained one large network corresponding to the highest level network in the pyramid. Figure~\ref{fig:traintime} shows that the Pyramid CNN method is able to achieve significantly better performance given the same amount of time. Training multiple networks in the lower scale levels also improves performance at the expense of moderate slow down.

\subsection{Results on Social Face Dataset}

As the room for further improvement of the verification accuracy on LFW becomes increasingly small, we propose another protocol to evaluate a face recognition system that reflects the requirements of real world applications. In many scenarios, especially in access control systems, the number of potential unmatched pairs is significantly more than the number of matched pairs. Though in normal cases most of the pictures received by the face verification system actually belongs to the authorized person, it is important that the false positive rate is small enough to stop the potential attackers from attempting the attack. In the current evaluation protocol where the accuracy is calculated on the point that the FPR and FNR are comparable, it is possible that an attacker will conduct one successful attack within reasonable number of attempts.

\setlength{\tabcolsep}{4pt}
\begin{table}
\begin{center}
\caption{TPR value on SFD. The LBP baseline is based on the high-dimensional LBP feature. It can be seen that there is considerable gap between the performance of our method and the baseline, and the gap is more significant on lower FRP threshold.}
\label{tab:jiayuan}
\begin{tabular}{lccc}
\hline\noalign{\smallskip}
 & LBP & Our Method & improvement\\
\noalign{\smallskip}
\hline
\noalign{\smallskip}
TPR@FPR=0.1 & 0.78  &  0.84 & 0.07\\
TPR@FPR=0.01  & 0.57 & 0.66 & 0.09\\
TPR@FPR=0.001  & 0.32 &  0.44 & 0.12\\
\hline
\end{tabular}
\end{center}
\end{table}
\setlength{\tabcolsep}{1.4pt}

We propose to measure the performance of the recognition system based on the point on ROC curve where the FPR is 0.001. This threshold suffices for most non-critical applications. Accurate measuring of the TPR value demands large number of face pairs. We suggest that the decision threshold should be computed based on at least one million unmatched face pairs and the threshold is used to test all matched face pairs to obtain the TPR value. In this experiment we gathered another dataset called Social Face Dataset which contains 48927 faces that rarely overlap with our training set. As shown in Table~\ref{tab:jiayuan}, there is significant gap between our method and the LBP baseline although the accuracy values on LFW are similar. We stress that to enable real-world access control applications, considerable effort is still required to improve the recognition system.

\section{Discussion}

We suggest that our Pyramid CNN can also be applied to other areas besides face. In object classification, even greater input resolution and the number of scale levels are typically used. Thus, the speed up of the Pyramid CNN should be more significant. However, there are challenges in the application of this technique. The face is a highly structured image with relatively fixed configuration of the parts. This is in contrast to the case of object classification or detection where the position of the main object has great variation. But we believe this technique can be extended to overcome the obstacle.

\end{document}